\pdfoutput=1
\documentclass[letterpaper, 10 pt, conference]{ieeeconf}  % Comment this line out if you need a4paper

\IEEEoverridecommandlockouts                              % This command is only needed if 
\usepackage{url}
\usepackage{graphicx}
\usepackage{booktabs}
\usepackage{romannum}
\usepackage{amsmath}
\usepackage{amsfonts}
\usepackage{multirow}
\usepackage{array}
\usepackage{cleveref}

\usepackage{algorithm} 
\usepackage{algpseudocode} 
\usepackage{longtable}
\newcommand{\squeezeup}{\vspace{-2.5mm}}
\UseRawInputEncoding

\title{\LARGE \bf
Proactive Action Visual Residual Reinforcement Learning for Contact-Rich Tasks Using a Torque-Controlled Robot}

\author{Yunlei Shi$^{1,2}$, Zhaopeng Chen$^{2,1}$， Hongxu Liu$^{3}$, Sebastian Riedel$^{2}$,\\ Chunhui Gao$^{2}$, Qian Feng$^{3}$, Jun Deng$^{2}$, Jianwei Zhang$^{1}$% <-this % stops a space
\thanks{*This research has received funding from the German Research Foundation (DFG) and the National Science Foundation of China (NSFC) in project Crossmodal Learning, DFG TRR-169/NSFC 61621136008, partially supported by European projects H2020 STEP2DYNA (691154) and ULTRACEPT (778602).}% <-this % stops a space
\thanks{{$^{1}$TAMS (Technical Aspects of Multimodal Systems), Department of
Informatics, Universität Hamburg}, {$^{2}$Agile Robots AG}, {$^{3}$Technische Universität München.}}}

\begin{document}
\maketitle
\thispagestyle{empty}
\pagestyle{empty}

%%%%%%%%%%%%%%%%%%%%%%%%%%%%%%%%%%%%%%%%%%%%%%%%%%%%%%%%%%%%%%%%%%%%%%%%%%%%%%%%
\begin{abstract}

Contact-rich manipulation tasks are commonly found in modern manufacturing settings. However, manually designing a robot controller is considered hard for traditional control methods as the controller requires an effective combination of modalities and vastly different characteristics. In this paper, we firstly consider incorporating operational space visual and haptic information into reinforcement learning(RL) methods to solve the target uncertainty problem in unstructured environments. Moreover, we propose a novel idea of introducing a proactive action to solve the partially observable Markov decision process problem. Together with these two ideas, our method can either adapt to reasonable variations in unstructured environments and improve the sample efficiency of policy learning. We evaluated our method on a task that involved inserting a random-access memory using a torque-controlled robot, and we tested the success rates of the different baselines used in the traditional methods. We proved that our method is robust and can tolerate environmental variations very well.

\end{abstract}

%%%%%%%%%%%%%%%%%%%%%%%%%%%%%%%%%%%%%%%%%%%%%%%%%%%%%%%%%%%%%%%%%%%%%%%%%%%%%%%%
\section{INTRODUCTION}
%\IEEEPARstart 
For high-precision assembly tasks, the robot needs to combine the high positioning accuracy with high flexibility. Designing a robot for these tasks is very challenging although such tasks can be easily performed by humans. Several torque-controlled robots have been designed for cooperative tasks to be performed in industrial environments \cite{albu2007dlr}, \cite{gaz2019dynamic}. These torque-controlled robots have seven revolute joints with torque sensors, and similar control algorithms \cite{albu2004passivity}, \cite{ott2004passivity}, \cite{albu2007unified}. Currently, torque-controlled robots are already safe enough when collisions occur with environments or humans \cite{albu2007dlr}, \cite{haddadin2015robot}. However, their effectiveness in real-life and production scenarios is still not satisfactory.
% small and medium-sized enterprises (SMEs), Computers, communication, and consumer electronics (3C) products are updated very frequently. 
 
Torque-controlled robots often serve computers, communication, and consumer electronics (3C) product lines, which usually involve small but complex assembly tasks,  and need to be adjusted quickly and frequently. Nowadays there are a few 3C assemble factory lines \cite{youtube2} but they need a long time to build and setup in high precise, which are not suitable for small and medium-sized enterprises (SMEs) who have automation needs but cannot afford to upgrade the entire production line.
\begin{figure}[htbp]
\centerline{\includegraphics[width=8.8cm, height=4.3cm]{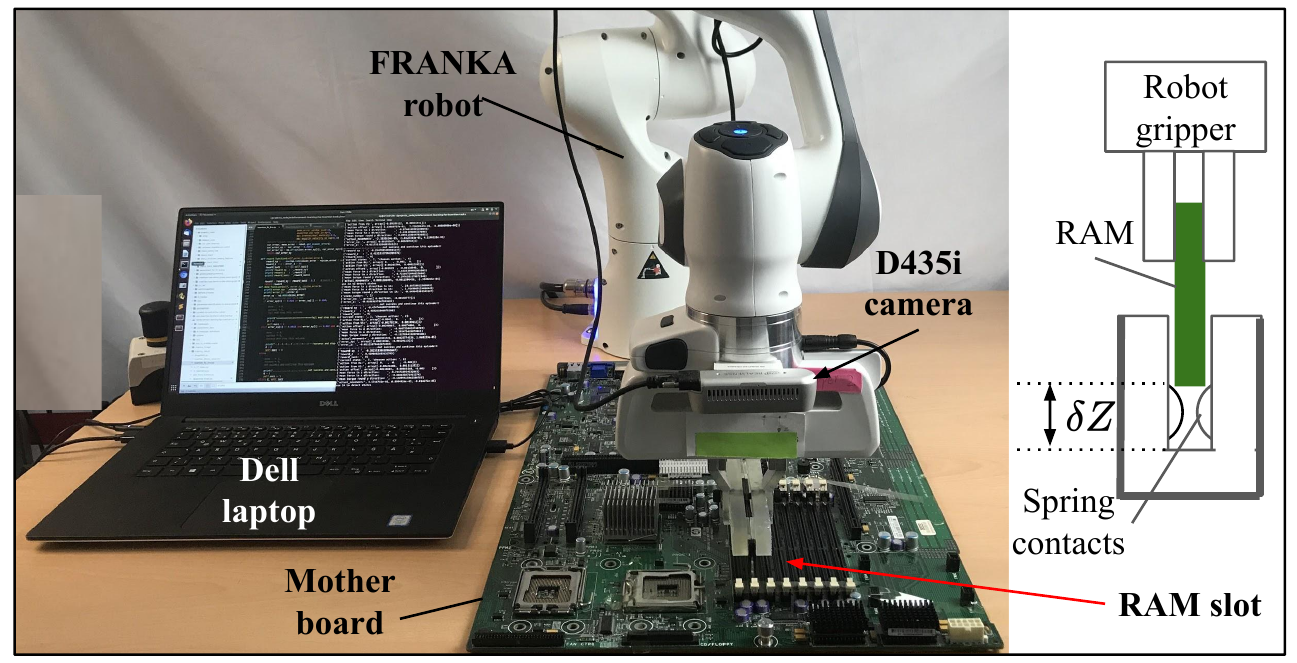}}
\caption{A contact-Rich task scenario: RAM insertion. Such kind of tasks always have stuck problems due to the tight clearance and narrow space.}
\label{experimentSetup}
\squeezeup
\squeezeup
\end{figure}Position uncertainty are quite normal in human-based traditional production lines. Some studies used simple fixed curves for exploring \cite{park2013intuitive}, \cite{youtube1} but they have low robustness against positional and angular errors for insertion tasks especially when targets are not fixed accurately. Schimmels and Peshkin \cite{peshkin1990programmed}, \cite{schimmels1992admittance} designed an admittance matrix for force-guided assembly in the absence of friction and after two years they improved the admittance control law. However, there still existed a maximum limit requirement of friction value \cite{schimmels1994force}. Stemmer et al. \cite{stemmer2006robust} proposed the region of attraction (ROA) method using vision and force perception to assemble the specified-shape objects, while geometry of the parts are required.

% \footnote{https://www.franka.de/capability/}
% \begin{figure}[htbp]
% \centerline{\includegraphics[width=8.5cm, height=2.8cm]{4robots.pdf}}
% \caption{From left to right: LWR-III from German Aerospace Center (DLR), KUKA LBR iiwa (intelligent industrial work assistant), FRANKA EMIKA, DIANA robot from AGILE ROBOTS AG.}
% \label{fig:4robots}
% \end{figure}

In this paper we equip a robot with a visual residual policy that combine multimodal feedback from vision and touch, two modalities with different frequencies, and characteristics. Our primary contributions are:

\noindent1) We propose a visual reinforcement learning (RL) method by combining a visual-based fixed policy with a contact-based parametric policy, this method greatly enhances the robustness and efficiency of the reinforcement learning (RL) algorithm. %\cite{johannink2018residual}, \cite{schoettler2019deep}
%use robot joint torque and visual sensory data, as well as robot proprioceptive data combined with  to solve the last-inch insertion problem  .

\noindent2) We propose the proactive action in the visual residual RL policy to solve the partially observable Markov decision process (POMDP) problem, which could ensure the task success rate and the ability to tolerate environmental variations.

\noindent3) We implement ablative and comparative studies to give the effects of each modality on task success rate and prove the robustness of our method in experiment.

% demonstrate insertion tasks in real industrial scenarios,  give the effects of each modality on task success rate.

\section{ BACKGROUND AND RELATED WORK} \label{BACKGROUND AND RELATED WORK}

% \subsection{3C Manufacturing Application of Robots}
% There are typically four steps in 3C manufacturing: part production, assembly, inspection, and packaging.  Robots always used in last 3 steps and assembly remains the most difficulty steps.

% Usually the procedure of assembly manipulation was divided into gross motion planning and fine motion planning. The former involves reasoning about the manipulator motion on the macro scale and considers its overall global movement strategy, while the latter considers how to deal with uncertainty to accomplish a task requiring high precision robustly \cite{siciliano2016springer}. In this paper, we focus on the robust algorithm of high precision assemble tasks.

% Now days there are a few PC assemble factory lines \cite{youtube2} but need complex setup with long time and accurate teach or image matching, or some demos/experiments use simple fixed curve for exploring \cite{youtube2}, \cite{park2013intuitive} but have low robustness to against positional and angular errors for a inserting task.
\subsection{Torque-controlled Robot Concepts}\label{Torque-controlled Robot Concepts}
Torque-controlled robots have been developed for unstructured environments that are fundamentally different from the environments where classical industrial robotics have been used. 
% \begin{figure}[htbp]
% \centerline{\includegraphics[width=9cm, height=3cm]{4robots.jpg}}
% \caption{From left: the DLR LWR-III (DLR), the KUKA LBR iiwa, FRANKA EMIKA (FRANKA EMIKA GmbH), DIANA(Agile Robots AG)!!!need last confirm!!}
% \label{fig:4robots}
% \end{figure}
%, which would result in several millimeters of positional error in torque control mode.
% \begin{figure}[htbp]
% \centerline{\includegraphics[width=\columnwidth]{lowlevelcontroller.pdf}}
% \caption{Illustration of the robot’s low-level control scheme. The actions $x_{des}$ are computed at a low frequency, and the desired joint torques calculated directly by the Cartesian impedance controller at 1000 Hz. The joint controller runs the torque feedback controller at 3000 Hz}
% \label{fig:lowlevelcontroller}
% \squeezeup
% \squeezeup
% \end{figure}
% \begin{figure}[htbp]
% \centerline{\includegraphics[width=8.2cm, height=4.0cm]{pbvs.pdf}}
% \caption{Position-based visual servo (PBVS) structure. ${^c}{x}{_d}$ is the desired end-effector pose relative to a target, while  ${^c}{x}$ is the current estimated end-effector pose relative to a target.}
% \label{PBVS}
% \squeezeup
% \squeezeup
% \end{figure}
The torque sensor in each joint plays a key role in robot controller.
The basic controller consists of a torque feedback loop, which can be interpreted as the scaling of the motor inertia $B$ to the desired value $B_\theta$ \cite{ott2004passivity}:
\begin{equation}
\setlength{\abovedisplayskip}{3pt}
\setlength{\belowdisplayskip}{3pt}
   \boldsymbol{\tau}_m=\boldsymbol{B} \boldsymbol{B}_\theta ^{-1} \boldsymbol{\tau_u} +(\boldsymbol{I}-\boldsymbol{B} \boldsymbol{B}_\theta ^{-1})\boldsymbol{\tau}
  \label{torquecontrol}
\end{equation}
% , whereas the maximum torque sensor error is around 0.5\% \cite{albu2007dlr}
Here, $\tau_u$ is an intermediate control input that could shape the Cartesian or joint impedance behavior \cite{albu2004passivity}, and $\tau$ is the joint torque data measured by the torque sensor as well as the torque vector applied to manipulators’ joints. $\tau_m$ is the torque on demand of the motor controller.
For the Cartesian impedance behavior, we have
\begin{equation}\label{cartImpController}
\setlength{\abovedisplayskip}{3pt}
\setlength{\belowdisplayskip}{3pt}
\begin{split}
&\boldsymbol{\tau}_u = -\boldsymbol{J}(\boldsymbol{\theta})^T(\boldsymbol{K}_x\boldsymbol{\tilde{x}}(\boldsymbol{\theta})+\boldsymbol{D}_x\boldsymbol{\dot x}(\boldsymbol{\theta}))+\boldsymbol{\overline{g}}(\boldsymbol{\theta})\\
&\boldsymbol{\tilde{x}}(\boldsymbol{\theta})=f(\boldsymbol{\theta})-\boldsymbol{x}_s\\
&\boldsymbol{\dot{x}}(\boldsymbol{\theta})= \boldsymbol{J}(\boldsymbol{\theta})\boldsymbol{\dot\theta}
\end{split}
% \label{cartcontrol}
\end{equation}
% For the impedance behavior of joint coordinates, we have
% \begin{equation}
% \tau_u=-K_{\theta}(\theta-\theta_s)-D_{\theta}\dot\theta+\~g(\theta)
% \label{jointcontrol}
% \end{equation}
$\boldsymbol{K_x}$ and $\boldsymbol{D_x}$ are the permutation and diagonal matrices of desired stiffness and damping; $\boldsymbol{x_s}$ is the desired end-effector (EE) pose, and $\boldsymbol{x}(\boldsymbol{\theta}) = f(\boldsymbol{\theta})$ is the EE pose computed based on the motor position. $\boldsymbol{J}(\boldsymbol{\theta})=\partial f(\boldsymbol{\theta})/\boldsymbol{\theta}$ is the manipulator Jacobian; $\boldsymbol{\theta}$ and $\boldsymbol{\theta}_s$ are the measured and desired motor positions, respectively. $\boldsymbol{\overline{g}}(\boldsymbol{\theta})$ is the gravity function that always comes from the CAD model or the parameter identification; this function inevitably has errors.

% Equations (\ref{torquecontrol}) and (\ref{cartImpController}) are the basic control laws of torque-controlled robots proposed by the German Aerospace Center (DLR). Usually, the maximum joint stiffness $\boldsymbol{K}_\theta$ can reach 4000 Nm/rad whereas the maximum Cartesian stiffness $\boldsymbol{K}_x$ can reach 4000 N/m \cite{albu2007unified}. The controller cannot add an integration part due to the spring-damper design concepts, and the upper stiffness is limited by joint flexibility.

% Based on the torque control concept \cite{ott2004passivity}, \cite{albu2004passivity}, Chen et al. \cite{chen2012task}, \cite{chen2010experimental} performed studies of joint/Cartesian impedance manipulation using the dexterous robotic hand DLR-HIT II. We implemented the Cartesian impedance motion controller in the explore phase as \Cref{fig:lowlevelcontroller}.

\subsection{Visual Servo Control in Manufacturing Application} \label{Visual Servo Control in Manufacturing Application}
Vision sensor allows robot to measure the environment with noncontact method. Shirai and Inoue \cite{shirai1973guiding} described an idea on how to use visual feedback to correct the position of a robot in order to increase assembly task accuracy.
% considerable effort has been devoted to the visual control of robot manipulators.
Position-based visual servo (PBVS) systems and image-based visual servo (IBVS) systems are two major classes of visual servo control systems. The typical control structure of PBVS can be found in \cite{hutchinson1996tutorial}.

An end-effector mounted camera could acquire the target depth and orientation information which can be used directly for PBVS \cite{teuliere2012direct}, \cite{fujimoto2003visual}. 
% Mount the small visual sensor on the robot
% Visual servoing is the fusion of image processing, kinematics, dynamics and control theory.
While the lens and the imaging sensors, the calibration of intrinsic/extrinsic parameters, the reflection, shadow and occlusion will exert a strong influence on the precision of the visual guidance \cite{li2019survey}.
\subsection{Reinforcement Learning for Assembly Tasks}
RL offers a set of tools for the design of sophisticated robotic behaviors that are difficult to engineer. RL has been applied previously and has gained great success in solving a variety of problems in robotic manipulations \cite{lee2018making}, \cite{luo2018deep}, \cite{schoettler2019deep}, \cite{inoue2017deep}, \cite{luo2019reinforcement}. 
Newman et al. \cite{newman2001interpretation} inverted the mapping from the relative positions to the observed moments and trained the neural net to guide the robotic assembly. Inoue et al. \cite{inoue2017deep} used long short-term memory to learn algorithms with two threads (an action thread and a learning thread) for searing and inserting a peg into a tight hole; however, their methods required several pre-defined heuristics, flat searching surfaces, and also a long training time.

% Rajeswaran \cite{rajeswaran2017learning} and Vecerik et al \cite{vevcerik2017leveraging} developed the methods with multiple demonstrations to improve the policy throughout the learning procedure. Multiple demonstrations are difficult to collect in high-precision assembly tasks.

Residual RL could take advantage of the efficiency of conventional controllers and the flexibility of RL \cite{johannink2018residual}. The idea is to try to inject prior information into an RL algorithm to speed up the training process instead of randomly exploring from scratch. 
% In many assembly tasks,  conventional controllers could optimize the priory of environment interactions whereas RL could learn fine-grained hand-engineered feedback controllers using friction and contacts.
\begin{figure}[htbp]
\centerline{\includegraphics[width=\columnwidth]{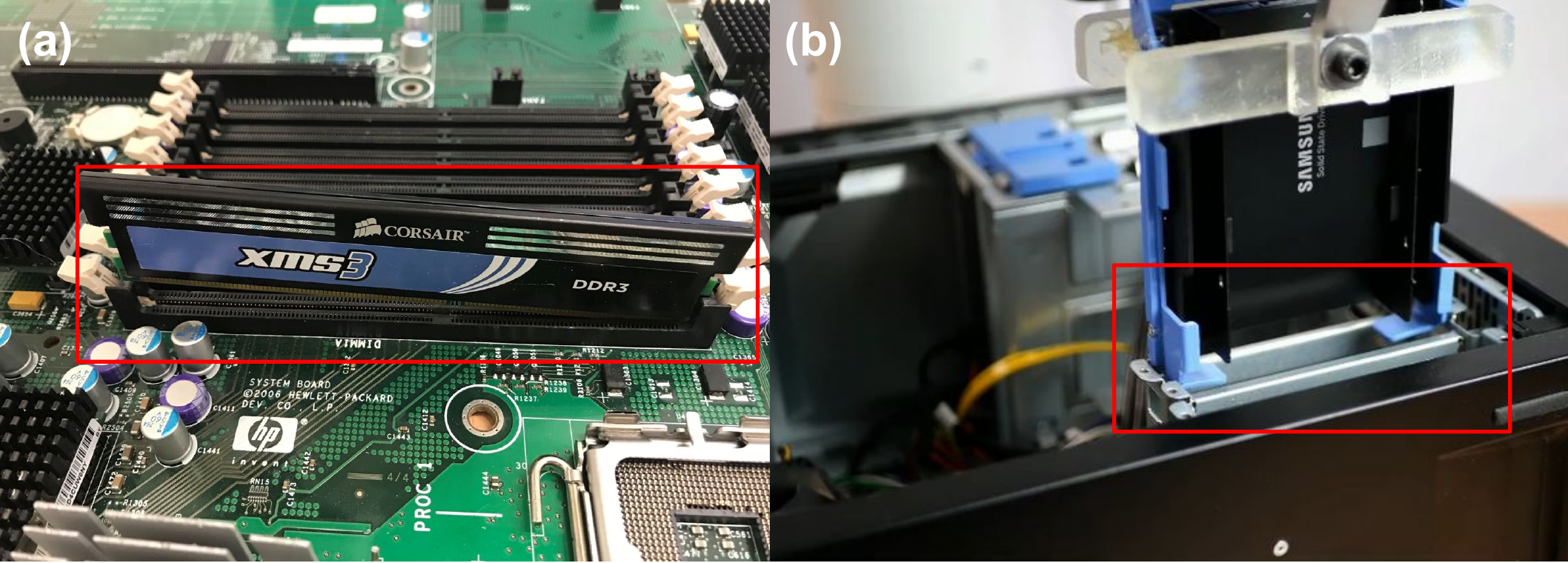}}
\caption{Inside the computer host, there is no sliding surface for insertion tasks. (a) RAM slot. (b) Solid State Disk (SSD) slot.}
\label{fig:nosliding}
\squeezeup
\end{figure}Specifying goals via images makes it possible to specify goals with minimal manual effort such as taking a photo \cite{schoettler2019deep}. Combining the sense of vision and touch could endow robots with a similar ability as humans to complete the assembly tasks \cite{lee2018making}, which could provide robustness to sensor and actuator noises \cite{schoettler2019deep} as well as position uncertainty. However, only a few studies have focused on real industrial production contact-rich tasks, and they also require a sliding surface for the algorithms to search \cite{lee2018making}, \cite{inoue2017deep}, \cite{lee2020guided}.
\section{PROBLEM STATEMENT AND METHOD OVERVIEW}
\subsection{Problem Statement}
\subsubsection{Position Uncertainty in Unstructured Environments}\label{Target Uncertainty in Unstructured Environments}

As we talked in \Cref{Torque-controlled Robot Concepts}, position uncertainty are quite normal in human based production lines. Workers could perform high-precision robotic assembly tasks with  their strong intelligence, excellent visual ability, and dexterous hands. while these tasks are very challenging to robots especially in these unstructured production environments.

% Torque-controlled robots have lower repeatability compared with position-controlled robot. For example, a 7-DoF torque-controlled robot has 0.1 mm pose repeatability\footnote{https://s3-eu-central-1.amazonaws.com/franka-de-uploads/uploads/Datasheet-EN.pdf} while 6-DoF position-controlled robot could reach 0.01mm.

Also, torque-controlled robots have low position robustness to friction and obstruction in contact-rich tasks due to the low stiffness design concepts as we described in \Cref{Torque-controlled Robot Concepts}. The limited control stiffness together with the friction and obstruction in contact-rich tasks give the position control error at the millimeter level. Torque-controlled robots are expected to achieve a desired dynamical relationship between environmental forces and movements of the robot in order to avoid breaking the environments or targets, thus the desired position and contact force can not be satisfied in the same dimension simultaneously.
Moreover, the location of the targets is uncertain sometimes due to insufficient accuracy of industrial assembly line.
% ($K_\theta$ or $K_x$)

Using visual method to correct the positions of the targets is an intuitive solution, while we still have position control problems when robot contact with targets due to the reason as we explained in \cref{Visual Servo Control in Manufacturing Application}, even we have implemented some explore actions (e.g., the spiral explore method \cite{park2013intuitive}).

% The latest generation of lightweight robots from DLR called safe autonomous robotic assistant (SARA)\footnote{https://www.dlr.de/rm/desktopdefault.aspx/tabid-11709/#gallery/29681} could solve the upright accuracy problem through inertial measurement units installed in base and EE, maximum stiffness was increased to 20000 N/m, but inaccurate position problem will still exist when load parameters are not accurate.
% ; this influence increased with the increase in the payload, and the environment
In 3C production lines, the insertion scenarios are different with the typical simplification settings of peg-in-hole \cite{lee2018making}, \cite{inoue2017deep}. For example, the random-access memory (RAM) insert-type task has problems as follow:

\noindent 1) The RAM slot or other slots does not have a proper surface for sliding behavior of the robot in alignment stage \cite{lee2018making}, \cite{luo2018deep} as shown in \Cref{fig:nosliding} which makes sliding type algorithm not work any more.

\noindent 2) The objects (like the RAM or hard disk) would be easily stuck by the structure near the slot or the slot itself in the explore/alignment stage as shown in \Cref{fig:nosliding}.

\noindent 3) Compared with previous studies, the slot has a long and narrow shape with tight clearance which is hard to insert by random and traditional search algorithm \cite{park2013intuitive}, \cite{park2017compliance}.
\subsubsection{Uncertainty POMDP States}
The main challenge of the traditional policy is to design adaptable, yet robust algorithms in the face of inherent difficulties for modeling all possible interaction behaviors. RL enabled us to find new control policies automatically for contact-rich problems where traditional heuristics had been used, but the results were not satisfactory.

Contact states are hard to estimate due to the sensor noise and robot modeling error, changing the Markov decision process (MDP) to POMDP, which makes it significantly harder to find an optimal policy \cite{ng2003shaping}, and it also requires more training time. Belief state tracking is one way to deal with the POMDP problem \cite{littman1995efficient}, \cite{white1991survey}, \cite{lovejoy1991survey}, but this method takes too much time to find an optimal policy.
% Other heuristics methods such as \cite{cassandra1998exact} have also been proposed, for instance, the state was estimated from $\hat s=arg\;max_sp(s)$, and the action could be chosen according to $arg\;max_aQ^*(\hat s,a)$. While we hope to use proactive action to identify the POMDP states rather than respond to them after they have occurred. 
% In these methods, it is always difficult to give general guarantees. 

%\squeezeup
% \subsection{Method Overview}\label{Method Overview}
% \begin{figure}[htbp]
% \centerline{\includegraphics[width=8.5cm, height=7.0cm]{imagefeatureanddepth.pdf}}
% \caption{ Feature-based image-matching approach. (a1) Current image features. (b1) Teach image features. (a2) Current depth image. (b2) Teach depth image. 4 images are recorded by Intel RealSense D435i Camera. Black pixels in a(2) and b(2) crosspond to unavailable depth values.}
% \label{fig:imagematch}
% \squeezeup
% \squeezeup
% \end{figure}
\subsection{Method Overview}\label{Method Overview}
\begin{figure}[htbp]
\centerline{\includegraphics[width=8.2cm, height=7.5cm]{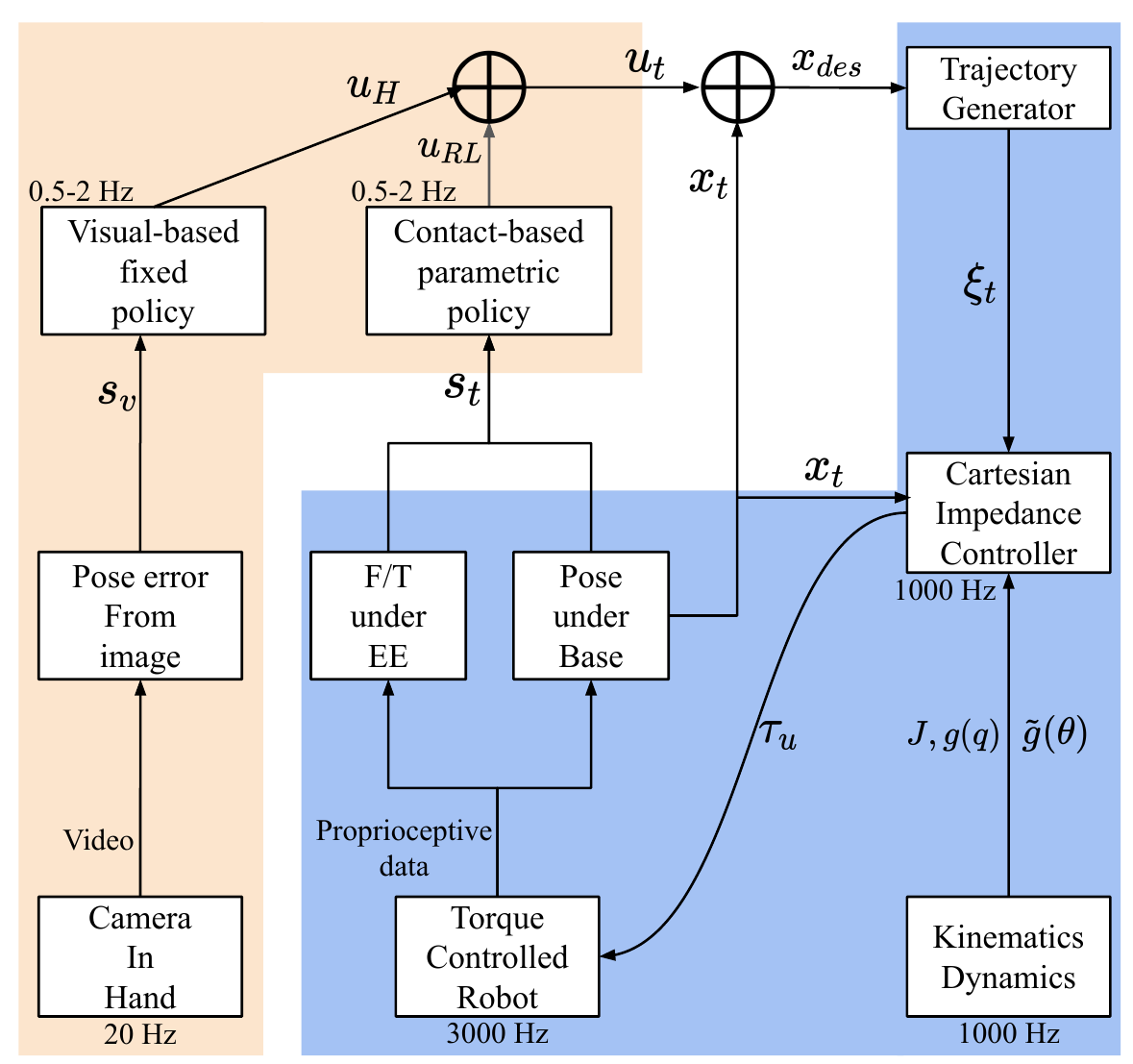}}
\caption{Representation of policies and controller scheme. Blue region part is the real-time controller, and the wheat region part is non-real-time trained policy.}
\label{fig:controlscheme}
\squeezeup
\squeezeup
\end{figure}

An eye-in-hand camera is helpful for solving the problem of position uncertainty in unstructured Environments in contact-rich tasks. The camera could try to align the characters of the target and compensate for the position error of the robot. Visual feedback control could provide geometric object properties for the pre-reaching target phase, whereas the camera aligning accuracy would always be disturbed by the target material or light. Force feedback control is quite helpful for providing contact information between the object and environment for accurate localization and control under occlusions or bad vision conditions, and force information could be obtained easily from the proprioceptive data in the torque-controlled robot controller.
% as shown in \Cref{fig:imagematch}.%
Visual feedback and force feedback are complementary and sometimes concurrent during contact-rich manipulation. 
In this paper, we implemented the visual-based fixed policy combined with contact-based parametric policy (see \Cref{fig:controlscheme}) as follow:

\noindent 1) For roughly locating the slot, we use one global image take from the teach mode with the RGB-D camera and rely only on the PBVS method \cite{hutchinson1996tutorial} (i.e. the visual-based fixed policy) control in this phase, because in free space, the contact-based parametric policy can not receive proper contact information.

\noindent 2) After rough location phase finished, the robot will move to target slot according to the pre-recorded transformation $^g\boldsymbol{x}_d$ from global image pose to detailed image pose, where $^g\boldsymbol{x}_d$ is recorded in teach phase. When the RAM in EE contact with the target slot, the detailed image which more accuracy for locating a slot, will be used to insert the RAM into the slot according to our method descirbed in \Cref{POLICY AND CONTROLLER DESIGN}.

% \noindent 3) We use the pose error between the current picture and target picture as well as the depth information as the RL reward function input.
\section{POLICY AND CONTROLLER DESIGN}\label{POLICY AND CONTROLLER DESIGN}

\subsection{Policy Design}\label{policydesign}

\subsubsection{\textbf{Visual Residual Reinforcement Learning}}

To take advantage of the high flexibility of RL and the high efficiency of conventional controllers, we introduce the idea of residual RL from \cite{johannink2018residual} with vision information; 
our method is expected to perform better compared with original residual RL in a variable environment due to the position uncertainty  problem in \Cref{Target Uncertainty in Unstructured Environments}. 

In residual RL, the policy are chosen by additively combining a fixed policy $\pi_H(s_v)$ with a parametric RL policy $\pi(\theta)(u_t|s_t)$: $u_t=\pi_H(s_v) + \pi_\theta(s_t)$. The fixed policy can help the agent move to the target, but prevent the agent from exploring more states. To balance the exploration and exploitation between the fixed policy and parametric RL policy, we design the weighted residual RL as 
\begin{equation}
\setlength{\abovedisplayskip}{3pt}
\setlength{\belowdisplayskip}{3pt}
  u_t =(1-\alpha) \pi_H(s_v) + \alpha * \pi_\theta(s_t).
  \label{Vision Residual Reinforcement Learning}
\end{equation}
Here, $\alpha$ is the action weight between the fixed policy and the parametric RL policy; The parametric policy is learned in the RL process to maximize expected returns on the task. We use a P-controller as the hand-designed controller $\pi_H(s_v)$ in our experiments for the visual-based fixed policy. 

Firstly, we explain the detailed design of $\pi_H(s_v)$. $s_v$ represents a geometric relationship of robot states which is a Euclidean distance calculated by visual and estimated depth information. % as shown in \Cref{fig:imagematch}. 
We introduce the method from \cite{martinet1996vision} which used depth information in PBVS. Combined with feature extraction and features' depth information $Z_N$, we could get estimated target feature set $^cP^*=(X^*_1, Y^*_1, Z^*_1, ..., X^*_N, Y^*_N, Z^*_N)$ and current feature set $^cP=(X_1, Y_1, Z_1, ..., X_N, Y_N, Z_N)$  whose coordinates are expressed with respect to the camera coordinate frame $c$ following the perspective projection method \cite{hutchinson1996tutorial}:
\begin{figure}[htbp]
\centerline{\includegraphics[width=8cm,height=4.1cm]{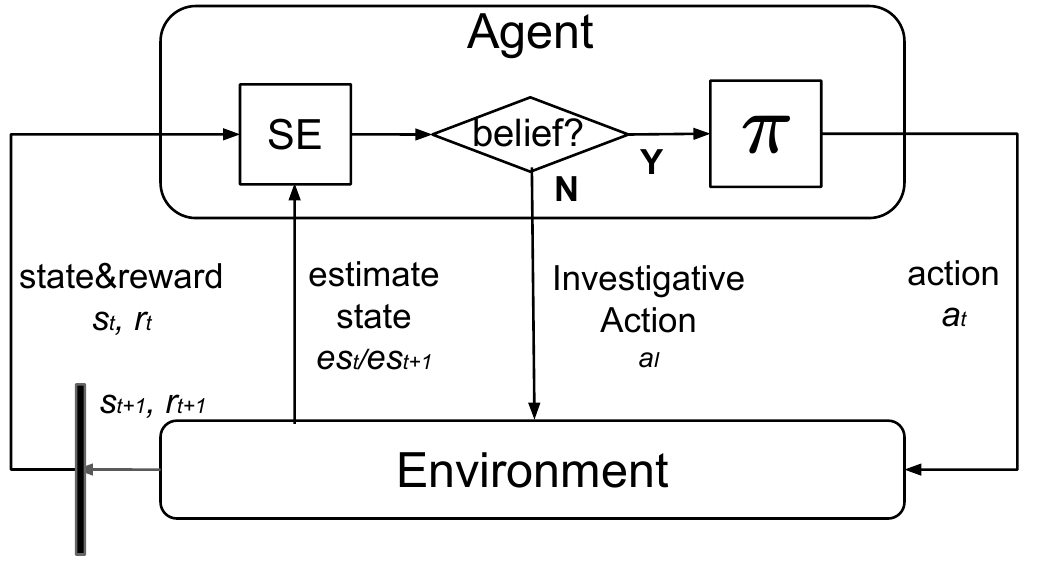}}
\caption{Investigative action idea for solving POMDP problem. SE: state estimator. The states will be estimated by SE function, then the policy receive the clear states and output an action.}
\label{IAMDP}
\squeezeup
\squeezeup
\end{figure}
\begin{equation}
\setlength{\abovedisplayskip}{3pt}
\setlength{\belowdisplayskip}{3pt}
  \begin{bmatrix} X_N \\ Y_N \end{bmatrix}=\frac{Z_N}{f}\begin{bmatrix} u_N \\ v_N \end{bmatrix}. 
  \label{perspective projection}
\end{equation}
Where $f$ is the focal length of the camera lens. $[u_N, v_N]^T$ gives the coordinates of the image feature set expressed in pixel units. Iterative Closest Point (ICP) \cite{besl1992method} could be used to get the coordinate transformation  ${^{c*}}\boldsymbol{x}_c$ by the feature set $^c\boldsymbol{P}$ and $^c\boldsymbol{P}^*$. 
\begin{equation}
\setlength{\abovedisplayskip}{3pt}
\setlength{\belowdisplayskip}{3pt}
\begin{split}
  &{^{c*}}\boldsymbol{x}_c=  
   \begin{pmatrix}
   {^{c*}}\boldsymbol{R}_c & {^{c*}}\boldsymbol{t}_c \\
   \boldsymbol{0} & 1
  \end{pmatrix} \\
\end{split}  
\label{transformationerror}
\end{equation}
Here we set $s_v=(^{c\ast}\boldsymbol{t}_c, \theta  \boldsymbol{u})$ depends on \Cref{transformationerror}, where $^{c\ast}\boldsymbol{t}_c$ is the translation error vector, and $\theta  \boldsymbol{u}$ gives the angle/axis representation for the rotation error \cite{siciliano2016springer}.
Then a velocity control scheme is designed by using an exponential and decoupled decrease of the error(i. e., $\boldsymbol{\dot e}=-\lambda \boldsymbol{e}$) as:
\begin{equation}\label{controlscheme}
\setlength{\abovedisplayskip}{3pt}
\setlength{\belowdisplayskip}{3pt}
\begin{split}
&\boldsymbol{v_c} = -\lambda (^{c*}\boldsymbol{R}_c)^T{^{c\ast}\boldsymbol{t}_c} \\
&\boldsymbol{w_c} = -\lambda \theta  \boldsymbol{u}
\end{split}
\end{equation}
\Cref{controlscheme} is used in rough location phase in \cref{Method Overview}. $[\boldsymbol{v}_c,\boldsymbol{w}_c]^T$ is the camera frame velocity command under current camera frame $\mathcal{F}_c$, which could be easily transfer to robot EE frame $\mathcal{F}_e$. In this paper, we calculate robot movement commands under robot EE frame $\mathcal{F}_e$ first, and then transfer them to base frame before sending to \Cref{cartImpController}.   
Secondly, we directly use $s_v=(^{c\ast}\boldsymbol{t}_c, \theta  \boldsymbol{u})$ as the states of fixed policy in accurate location phase, 
\begin{equation}\label{controlscheme2}
\setlength{\abovedisplayskip}{3pt}
\setlength{\belowdisplayskip}{3pt}
\pi_H(s_v) = -\boldsymbol{k}_p \cdot s_v,
\end{equation}
which is quite convenient to implement. 

In this paper, we use a value-based RL called Q-learning algorithm as the contact-based parametric RL policy $\pi(\theta)(u_t|s_t)$, the Q-function is implemented as a table with states as rows and actions as columns, then we can update the table by using the Bellman equation: 
\begin{equation}\label{BellmanEquation}
\setlength{\abovedisplayskip}{3pt}
\setlength{\belowdisplayskip}{3pt}
\begin{split}
Q^{\pi}(s_t,u_t)=\mathbb{E}_{r_t,s_{t+1}\sim E}[r_t+\gamma\mathbb{E}_{u_{t+1}\sim\pi}[Q^{\pi}(s_{t+1},u_{t+1})]].
\end{split}
\end{equation}

\subsubsection{\textbf{Proactive Action}}\label{Investigative Action}
Most studies \cite{johannink2018residual}, \cite{luo2019reinforcement}, and  \cite{lee2018making} have modeled the robot manipulation task as a finite-horizon discounted Markov Decision Process (MDP) $\mathcal{M}$ in an environment $E$, with a state space $\mathcal{S}$, an action space $\mathcal{A}$, state transition dynamics $\mathcal{T}$ : $\mathcal{S} \times \mathcal{A} \to  \mathcal{S}$, a discount factor $\gamma\in(0,1]$, and a reward function  $r: \mathcal{S}\times\mathcal{A} \to \mathcal{R}$ to determine an optimal stochastic policy $\pi$. 
\begin{figure}[htbp]
\centerline{\includegraphics[width=\columnwidth, height=4.4cm]{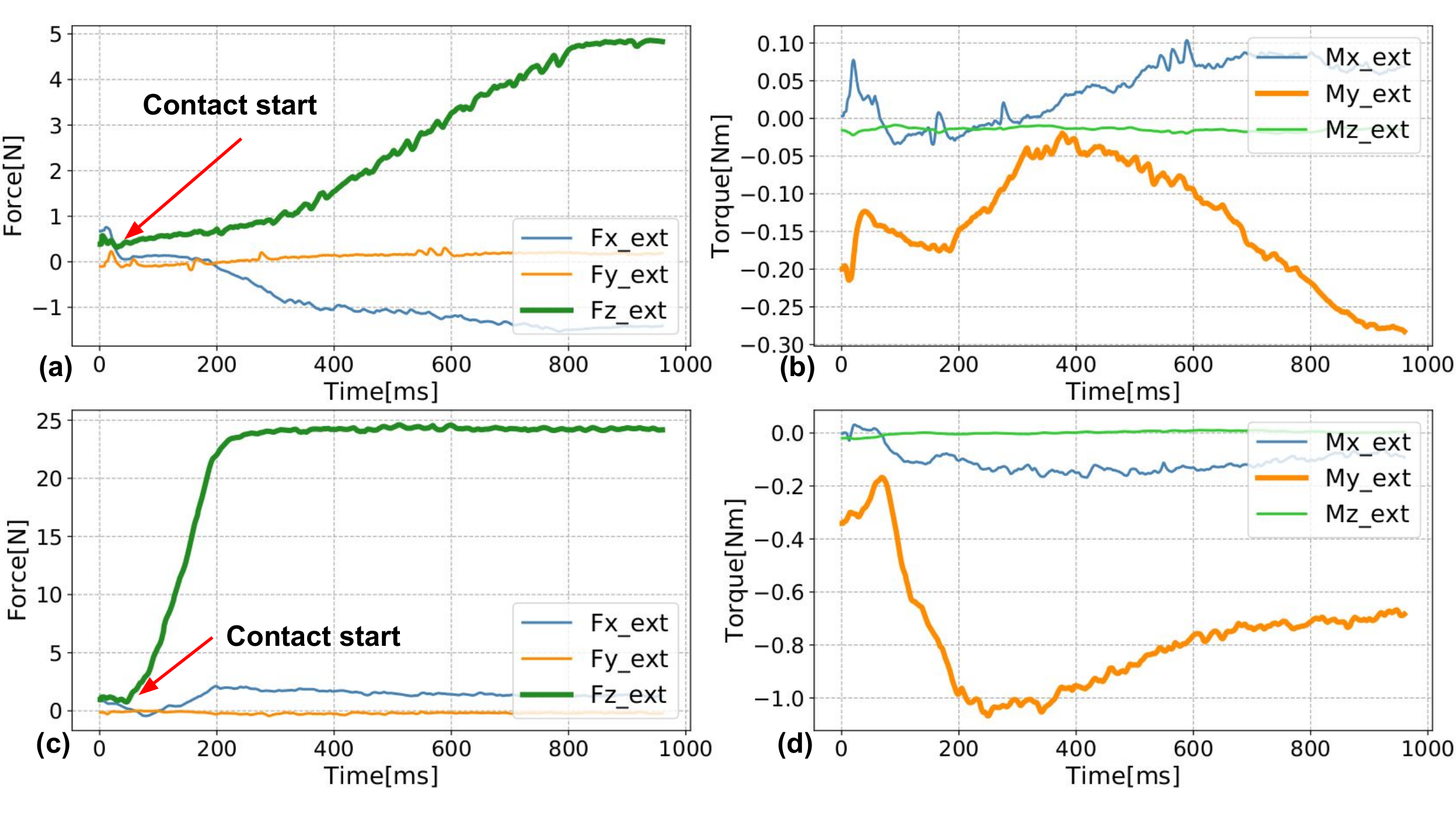}}
\caption{(a) RAM contacts with one slot side in \textbf{movement action} with 5 N force feedback in the Z direction. (b) External moment data $M_y$ which is hard to detect the torque contact status (goes up first and goes down later during contact force increase). (c) RAM contacts with the same side of (a) using an \textbf{investigative action} with 25 N press force. (d): External moment $M_y$ reaches to -1 Nm which could clearly detect contact status}
\label{FMcurve}
\squeezeup
\squeezeup
\end{figure}
In practice, many contact states $s_t$ cannot be observed directly in the manipulation tasks that are close to a POMDP problem. However, the POMDP problem is confined to the modeling error of the torque-controlled robot, which makes it difficult to detect the contact states.
Inspired by wild gorillas, who tried to cross a pool of water using a walking stick to test the water depth  \cite{breuer2005first}, we improved our RL process by adding a proactively investigative action ($a_I$) that could detect the clear states ($es_t$) involved in the RL process as \Cref{IAMDP} which is different with \cite{inoue2017deep} who continues push the target get a detectable moment; here, investigative action is one kind of the proactive actions.
% By doing this, our robotic contact-rich task can still be solved by using the MDP-based RL, which would be easier than directly solving a POMDP problem.

We use the investigative action $a_I$ combined with $u_t$ to construct a new policy $u^I_t(s_t)$ instead of the original $u_t(s_t)$, which can be written as $a_I,u_t \to \textit{E} \to s^I_{t+1} $. Where $s^I_{t+1}$ is determined by adding an investigative action $a_I$ of the torque-controlled robot to the environment. Consequently, the heuristic design of the investigative action prevents the learning process from falling into multiple unclear states. 
% This makes it effective and efficient in terms of the optimization.

In particular, the torque-controlled robot outputs either the movements or the forces. In our experiments, the movements are taken as the actions in the action space  \( \mathcal{A} \), and the forces are taken as the investigative actions.
Instead of using 20 N force continuously to detect the values of the moments in the search phase \cite{inoue2017deep},
% \begin{figure}[htbp]
% \centerline{\includegraphics[width=8.cm, height=3cm]{RAMslot.pdf}}
% \caption{Description of RAM insert task details. RAM always stuck by side A or B.}
% \label{RAMslot}
% \squeezeup
% \squeezeup
% \end{figure}
we only command the controller to exert a force (10-25 N) in some directions in a short time (0.5-1 s) as the investigative action while the feedback movements or force/moments are used to verify the contact states when the states are not clear. Our investigative action method can greatly reduce the friction and the probability of getting stuck when the robot performs the movement actions.
\subsection{Controller Design:}
We use the increment equation $x_{des}=x_t + u_t$ to avoid the potential ``far away" problem for safety concerns; $x_{des}$ is the desired EE pose, and $x_t$ is the current EE pose; $u_t$ is the increment action command from the agent.
\begin{algorithm}
	\caption{Visual Residual Reinforcement Learning with Investigative Action}
	\algorithmicrequire{\;RL policy $\pi_\theta$, fixed policy $\pi_H$}.
	\begin{algorithmic}[1]
		\For {iteration=1\;to\; M episodes}
		\State Copy latest policy $\pi_\theta$ from learning thread
		\State Sample initial state $s_0$
			\For {step=1\;to\; N}
				\State Get action $u_{RL}$ by greedily picking from $\pi_\theta(s_t)$
				\State Get action $u_{H}$ from $\pi_H(s_v)$
				\State Output policy action: $u_t =(1-\alpha) u_H + \alpha * u_{RL}$
                \If {belief ==true}
                    \State Get next state $u_t \to s_{t+1}$
                \Else {}
                    \State Get next state $a_I,u_t  \to s_{t+1} $ 
                \EndIf
				\State Optimize $\pi_\theta$ with Equation (\ref{BellmanEquation})
				\If {EpisodeEnd == true}
        		\State break
        		\EndIf
			\EndFor
		\EndFor
	\end{algorithmic} 
\end{algorithm}
The Cartesian impedance controller takes the Cartesian EE movement $u_t$ from agent at 0.5 to 2 Hz, and the output joint torque gives the command $\tau_u$ to the robot at 1000 Hz. We calculate the desired EE pose $x_{des}$ by combining $u_t$ with the current EE pose $x_t$. 
The trajectory generator bridges the low frequency output $x_{des}$ of the agent and the high frequency impedance control of the robot and outputs $\xi_t=x_s$ to the Cartesian impedance controller in \Cref{cartImpController}. $x_k$ is the position and $q_k$ is the quaternion representation of the orientation given by a simple linear interpolator:
\begin{equation}\label{controller}
\setlength{\abovedisplayskip}{3pt}
\setlength{\belowdisplayskip}{3pt}
\begin{split}
& \xi_t=\{\boldsymbol{x}_k,\boldsymbol{q}_k \}_{k=t} ^{t+T}.
\end{split}
% \squeezeup
\end{equation}
% & \~x(\theta) = f(\theta)-\xi_t \\
\section{EXPERIMENTS: DESIGN AND SETUP}
We consider the experiment for the insertion task here. The task can be described as moving the already-grasped parts to their goal pose as shown in \Cref{experimentSetup}. This is the most common setting in manufacturing. The success of such tasks can be measured by minimizing the distance between the objects and their goal pose especially in the Z direction (see \Cref{experimentSetup}).
% and Fig. \ref{RAMslot}
% Our experiments address on the following questions:

% \noindent 1) Can we learn a policy for insertion with generalization capabilities for the torque-controlled robot by considering the force and torque measurements?

% \noindent 2) Does the visual residual RL improve the performance and sample-efficiency of RL algorithms while decreasing the training time?

% \noindent 3) Can our method increase the robot's capability of tolerance of environmental variations?

% \begin{figure}[htbp]
% \centerline{\includegraphics[width=8.cm, height=8.cm]{experimentSetup.pdf}}
% \caption{(a) Description of RAM insert task details.RAM always stuck by side A or B. (b) Description of RAM insertion experiment components. }
% \label{experimentSetup}
% \end{figure}
\subsection{Experiment Algorithm Design}
In our weighted residual RL, actions $u_t$ are designed by adding the fixed policy $u_H = \pi_H(s_v)$ with the parametric policy $u_{RL} \sim \pi(\theta)(u_t|s_t)$:
\begin{equation}
\setlength{\abovedisplayskip}{3pt}
\setlength{\belowdisplayskip}{3pt}
  u_t =(1-\alpha) u_H + \alpha * u_{RL}.
  \label{weightRL}
% \squeezeup  
\end{equation}
The fixed policy output $u_H$ is calculated by a hand-designed controller as given in \Cref{controlscheme2}; $\alpha$ helps to adjust the balance between exploration and exploitation. We set $\boldsymbol{k}_p$ to (1,1,0.3,0,0,0) when the fixed policy is calculated. To identify a reasonable weight between the two components, we initially experimented with the weighted residual RL by introducing a group of action weight parameters, such as 0.3, 0.5, and 0.7. The training experiments suggested an optimum policy output with a weight of 0.5, whereas the weight could increase or decrease around 0.5 according to the visual condition in the implementation phase. We utilized the algorithm to detect states and implemented its slightly-modified version where the trained policies were constructed by two aforementioned components. Here the flag belief is set to 0 or 1, according to the moment threshold settings, a detectable moment(over threshold) always gives true belief state. Combined with the investigative action mentioned in Section \ref{Investigative Action}, the modified Q-learning algorithm was trained at a high speed, and it easily resulted in optimization. 
\subsubsection{Action Design}
We design Cartesian movement actions for this experiment. Each Cartesian movement dimension was set to +1 for a positive movement and -1 for a negative movement; therefore, we had $6*2=12$ actions. We set $\lambda$ as the scale parameter to adjust the amplitude of actions as
\begin{equation}
\setlength{\abovedisplayskip}{3pt}
\setlength{\belowdisplayskip}{3pt}
  \boldsymbol{a}=\lambda[P^d_{\sigma x}, P^d_{\sigma y}, P^d_{\sigma z}, R^d_{\sigma x}, R^d_{\sigma y}, R^d_{\sigma z}].
  \label{actions}
% \squeezeup 
\end{equation}
Here, $P$ and $R$ are positional and orientational movements under EE frame, respectively.  $\lambda$ is easy to choose because it is closely related to assembly clearance and visual accuracy, normally we set $\lambda=0.002$, then we have movements resolution at $0.002$ mm and $0.002$ rad level. We found that orientational movements accuracy were enough by using the fixed policy $u_H$, so we only output positional movements actions in our RL idea, this is normal setting because the visual feedback and force feedback are complementary during contact-rich manipulation.

The \textit{Investigative action} was designed as the force action $^{e}F_z= -25N$ under robot EE frame $\mathcal{F}_e$ for 1 s. The robot will try to add force but will stop moving if the force is greater than 25N or the movement is more than 3 mm. Then, the agent will obtain clear state feedback because of the large contact force and torque amplitude, as shown in \Cref{FMcurve}.

\subsubsection{Reward Design}
Depending on the pose error between the current picture and the target picture and the depth information, the reward function was set as follows:
\[
    r= 
\begin{cases}
    1,              & \text{(success)}\\
    -2,             & \text{(failed).}\\
    1 - 150 \Vert s_{xy} \Vert _2 - s/s_{max},& \text{(otherwise)}.

\end{cases}
\]
Here, $s_{xy}$ is the norm of the x and y errors of the images, $s$ is the number of steps in one episode, and $s_{max}$ is the maximum steps in one episode.
% \footnote{https://frankaemika.github.io/libfranka/index.html}, which is calculated by the proprioception torque sensors output
\subsubsection{State Design} 
We get the estimated 6-DoF external force and moments along the X, Y, and Z axis under the EE frame from Franka controller. Here we consider the contact force and the moments between the robot's EE (i.e., the grasping RAM) and the slot as the MDP states as
\begin{equation}
\setlength{\abovedisplayskip}{3pt}
\setlength{\belowdisplayskip}{3pt}
  \boldsymbol{s}=[F_x, F_y, F_z, M_x, M_y, M_z]
  \label{status}
\end{equation}
We assume that the EE contacts the slot when the external force $|F| > 4$ N or the external moments $|M| > 0.4$ Nm, a value of $\pm1$ means that a contact is made, and 0 means that there is no contact with the encoding states.
% We remove the unnecessary dimensions $[M_x,M_z]$ from the states according to \cite{inoue2017deep} to simplify the situation in this experiment.

\subsection{Experiment Environment and Task Setup}
\subsubsection{Environment Setup}
We used the Franka robot \cite{gaz2019dynamic} for real robot experiments and set the Cartesian stiffness as 3000 N/m and 300 Nm/rad (Recommended upper limit). Two sensor modalities were available in the real hardware, including proprioception and red–green–blue (RGB) depth camera. The RGB and depth information were recorded using the eye-in-hand Intel RealSense Depth Camera D435i. The policy ran on a Dell Precision 5510 laptop and sent the updated position to the real-time controller, which calculated the joint torque command and sent it to the robot controller at 1000 Hz. We used a CORSAIR DDR3 RAM and a motherboard as training and testing environment.
% \footnote{https://www.franka.de/technology}
% In our experiment, we started our policy from the slot at 3 cm and set random position errors in between at $\pm2 mm$ and $\pm3 mm$.
% We implemented our Proactive Action Visual RL method as \Cref{weightRL}.
\subsubsection{Tasks Setup}
In ablation study experiment, We evaluated our trained policy by masking different modalities as 4 baselines given below:

\noindent 1) \textit{No vision}: masks out the visual part action; we set $\alpha=1$.

\noindent 2) \textit{No RL policy}: masks out the RL part action; we set $\alpha=0$.

\noindent 3) \textit{Random policy}: generates a random Q table.

\noindent 4) \textit{No investigative action}: masks out the investigative action and chooses random action when the state is not clear.

We set maximum steps as 10 and add initial random errors($|error|\in[2,3]mm$) in x and y directions for each baseline only in ablation study experiment.

% because we concentrate on the ablation study in this setup)

% fixed and (we moved the motherboard around $\pm$5 cm and $\pm$15 degree) to 
In comparison study experiment, we compared the task success rates of our method with the other four baselines in the real scenarios (no maximum steps limit and no initial random errors for each baseline) by moving the motherboard, which are as follows:

\noindent 1) Baseline 1: For normal teaching and direct insertion

\noindent 2) Baseline 2: For normal teaching with spiral exploration

\noindent 3) Baseline 3: For teaching with vision and direct insertion

\noindent 4) Baseline 4: For teaching with vision and spiral exploration

\section{EXPERIMENTS: RESULTS AND DISCUSSION}

\begin{table}[t!]
	\caption{Ablation study of Policy evaluation statistics}
	\label{tab:Ablation experiments}
	\centering
	\begin{tabular}{lcc}
		\toprule
		\textbf{Baselines} & \textbf{Result(success/total)} & \textbf{Time Cost}\\
		\midrule
		No vision & 92/200 & 1.09 h \\
		No RL policy & 112/200 & 0.65 h \\
		Random RL policy & 77/200 & 2.59 h \\
		No investigative action & 66/200 & 0.85 h \\
	    \textbf{Our method}	 & \textbf{179/200} & 1.18 h \\
		\bottomrule
	\end{tabular}
\squeezeup
\squeezeup
\end{table}
We trained our policy with 500 episodes, and each episode lasted a maximum of 50 steps. The training time for the exploration was approximately 150 minutes which is much less than \cite{lee2018making}. We specified discrete actions in this experiment, and the action execution had errors. Our policy can increase the probability of success and decrease the cost steps but cannot guarantee success every time. We set random errors for the initial pose of the robot; sometimes, the robot will successfully insert by chance and obtain a high reward in the early stage of training. 
%  which is half time of \cite{lee2018making}

\Cref{tab:Ablation experiments} shows the ablation study result of the policy evaluation statistics.  \textit{Random RL policy} and \textit{No investigative action} had poor performances with success rates of 38.5\% and 33\%, respectively. \textit{No vision} had a 46\% success rate because of discrete overshooting actions whereas \textit{No RL policy} had a 56\% success rate because the RAM was always stuck by the short side of the slot. Our method had a success rate of 89.5\%. It should be noted that the success rate of our method is limited buy the maximum steps in the experiment.

We observed that the absence of either visual or correct forces/moments information negatively affected the task success rate, and wrong policy performance was even worse than without RL policy. Therefore, the \textit{Random RL policy} and \textit{No investigative action} had similar performances because the RL policy is always in conflict with the visual output action. None of the four baselines has reached the same level of performance as the final method. With visual input alone, the robot sometimes cannot overcome the last small distance because of either the limited movement accuracy of the robot or contact friction, while RL policy are capable of recovering from such issues whcih could be proved in our method. Without the visual input, the robot will require more steps to find the proper pose for insertion and will always overshoot for some actions (i.e., drop out of the slot).  
\newcommand{\tabincell}[2]{\begin{tabular}{@{}#1@{}}#2\end{tabular}}  
\begin{table}[t!]
	\caption{Comparison of success rates for different baselines}
	\label{tab:traditionalmethodbaselines}
	\centering
	\begin{tabular}{lcc}
		\toprule
		\textbf{\textbf{Baselines}} & {\textbf{Fix motherboard}} & {\textbf{Move motherboard}}\\
		\midrule
		Baseline 1 & 97/100  & 0/20 \\
		Baseline 2 & 100/100  & 0/20 \\
		Baseline 3 & 98/100  & 81/100\\
		Baseline 4 & 100/100  & 88/100 \\
	    \textbf{Our method}	& \textbf{100/100} & \textbf{100/100} \\
		\bottomrule
	\end{tabular}
\squeezeup
\squeezeup
\end{table}

\Cref{tab:traditionalmethodbaselines} shows a comparison of the success rates of different traditional method baselines. In order to simulate industrial scenario, the additional random error and maximum steps limit in ablation study are removed. Obviously, The baselines 1\&2 work well only when the motherboard are fixed in the same position as in the teaching phase, so we only test 20 times in ``move motherboard'' case for baselines 1\&2 for saving time. The success rates for the baselines 3\&4 are increased with vision correction, but they still have failure cases due to the visual error. Our method shows a strong ability to tolerate environmental variations and resilience from stuck with full success which really meets the requirements of industrial scenarios. Please note that in the comparison study, the increase of success rates is also related to the removal of initial errors and the removal of limit of the maximum steps.
% (i.e., the normal teach method)

%  The robot can not even go to right grasp position in RAM storage slot; however, this position error is similar to the insert position. Therefore, there could be only one success case.
% From the experiment, we also obtained some new ideas, such as setting up a dynamic $\alpha$ to obtain more actions from the visuals in the beginning and from the RL policy near the end stage; this would help decrease the time.

\section{CONCLUSION AND FUTURE WORK}

In this paper, we combined RL with an operational space visual controller to solve position uncertainty problem in high-precision assembly tasks, and we proposed a proactive action idea to solve the POMDP problem by using an investigative action.

% This proactive action idea could also be extended to other POMPD algorithms to predict and clarify the unclear states. 

Our method could solve the shortage of traditional visual servoing method by using our visual residual RL algorithm, we inherits some traditional controller parameters which makes setting up not fast enough, we will extend our method to be trained towards end-to-end approach in next step.

Unfortunately, space does not permit more generalize test in this paper, while we test the SSD insertion scenario as \Cref{fig:nosliding} with our policy and achieve full success with 100 times test. We will continue to generalize the model and policy so that they could handle different parts and robot manipulators in the next step. Then, the skill could be packaged as a service that will be delivered to robots in new factory lines with a short setup time.
Our method uses a discrete number of actions to perform the insertion task, as an obvious next step, we will analyze the difference between this method and continuous space learning techniques.
% , such as asynchronous advantage actor-critic (A3C) and deep deterministic policy gradient (DDPG) because we believe it is more similar to human's operation.
% Unlike most previous studies that use the FT sensor in the EE, we used only joint torque sensors to help train the policy which showed the robustness of our policy. 

% \clearpage
% \bibliographystyle{IEEEtran}
% \bibliography{bibliography}


\bibliographystyle{IEEEtran}
\bibliography{reference}

\end{document}